%% file: main.tex
\documentclass[letterpaper, 10 pt, conference]{ieeeconf}

\usepackage{adjustbox}
\usepackage{caption}
\usepackage{cite}

\usepackage{amsmath,amssymb,amsfonts}
\usepackage{graphicx}
\usepackage[linesnumbered,algoruled,boxed,lined]{algorithm2e}
\usepackage{textcomp}
\usepackage{comment}
\usepackage{xcolor}
\usepackage{paralist}
\def\BibTeX{{\rm B\kern-.05em{\sc i\kern-.025em b}\kern-.08em
    T\kern-.1667em\lower.7ex\hbox{E}\kern-.125emX}}
\usepackage{times}  
\usepackage{helvet}  
\usepackage{todonotes}
\usepackage{amsmath}
\usepackage{amssymb}
 
\usepackage{amsthm}
\usepackage{mathtools}
\usepackage{acronym}
\usepackage{courier}  
\input{defs.tex}
\usepackage{url}  
\usepackage{graphicx}  
\usepackage{bm}
\usepackage{color}
\usepackage{amsfonts}       
\usepackage{nicefrac}       
\usepackage{microtype}      
\usepackage{lipsum}
\usepackage{amsmath}
\usepackage{subcaption}
\usepackage{algpseudocode} 
\usepackage{tabularx}

\newsavebox{\measurebox}

\SetKwInput{KwInput}{Input}
\SetKwInput{KwOutput}{Output}

\title{\LARGE \bf Learning Contact-aware CPG-based Locomotion in a Soft Snake Robot
}

\author{Xuan Liu\textsuperscript{1}, Cagdas Onal\textsuperscript{1}, Jie Fu\textsuperscript{1}
\thanks{\textsuperscript{1}Xuan Liu is a PhD student under the supervision of Jie Fu and Cagdas Onal in the Robotics Engineering program at Worcester Polytechnic Institute.} 
\thanks{\tt\small xliu9, cdonal, jfu2@wpi.edu}%
\thanks{This work was supported in part by the National Science Foundation under grant \#1728412.}%
}

\begin{document}

\maketitle
\thispagestyle{empty}
\pagestyle{empty}

\begin{abstract}
In this paper, we present a model-free learning-based control scheme for the soft snake robot to improve its contact-aware locomotion performance in a cluttered environment. The control scheme includes two cooperative controllers: A bio-inspired controller (C1) that controls both the steering and velocity of the soft snake robot, and an event-triggered regulator (R2) that controls the steering of the snake in anticipation of obstacle contacts and during contact. The inputs from the two controllers are composed as the input to a Matsuoka CPG network to generate smooth and rhythmic actuation inputs to the soft snake. To enable stable and efficient learning with two controllers,  we develop a game-theoretic  process, fictitious play, to train C1 and R2 with a shared potential-field-based reward function for goal tracking tasks.  The proposed approach is tested and evaluated in the simulator and shows significant improvement of locomotion performance in the obstacle-based environment comparing to two baseline controllers.

\end{abstract}

\section{Introduction}
Soft continuum robots have unique advantages in traversing within cluttered and confined environments, due to their flexible body and deformable materials. Applications of soft continuum robots can include  search-and-rescue \cite{hawkes2017soft}, pipe inspection\cite{majidi2014soft} and medical surgery \cite{wang2017cable}.  In particular, comparing to the mobility of the legged animals who can only use their feet as push points, biological snakes have the unique advantage that any part of their body, if properly controlled, could benefit from the propulsion force generated by the contact with obstacles. In this paper, we investigate the following question: How to  design locomotion control for soft snakes that can intelligently adapt to the contact with the obstacles and even benefit from propulsion forces in an obstacle-based environment?

In literature, several research groups have studied  this unique feature   in snake robot locomotion. Transeth \etal \cite{transeth2008snake} first defined this feature as the \textit{obstacle-aided locomotion}, which means that the snake robot actively utilizes external objects other than flat ground to generate propulsion during the locomotion. In their pioneering work, they proposed a two-module task framework of the obstacle-aided locomotion with: (a) a path planner that looks for a trajectory with more active contact chance for the rigid snake robot, and (b) a motion controller that controls the snake robot's real-time body movements to optimally utilize the contacts between the robot and the environment and generate desired propulsion force for the locomotion. In \cite{liljeback2010hybrid}, a hybrid controller is developed, where a  contact event is detected and calculated to trigger a reactive control that maximizes the total propulsion force at the moment. This controller has been applied to a rigid snake robot and showed its reliability in maintaining beneficial propulsion force. When using this controller, a big challenge is that rigid snake robots can at times get jammed by obstacles. To handle this issue, Liljeback \etal \cite{liljeback2011experimental} proposed to detect a predefined jam event and resolve it by rotating the contacted links to increase the total propulsive contact force. Kano \etal \cite{kano2012local} proposed local reflexive mechanisms that interrogate the contact status between a soft-bodied scaffold snake robot and the obstacles to determine whether the contact is beneficial to the locomotion. However, the soft body in \cite{kano2012local} is not real but an approximation by increasing the degrees of freedom of each joint by using linear actuators. 


So far, the theoretical base of obstacle-aided locomotion\cite{7838565} is mainly studied and implemented with rigid snake robots. It is  appealing to enable such a capability for soft snakes due to the important tasks that are not feasible for rigid robots. However, few studies have investigated this control problem for soft snake robots. Moving from rigid snake robots to soft snake robots has a number of challenges: Firstly, the pneumatic actuators in soft snake robots have  nonlinear, delayed, and stochastic dynamical response given inputs, making it difficult to achieve fast responses comparing to rigid snake robot. Secondly, it is challenging to construct accurate kinematic/dynamic models for a soft snake robot, due to the continuum of the pneumatic actuators. Thus, the aforementioned model-based design will not be feasible for soft robots. 

\begin{figure}[h!]
    \centering
    \includegraphics[width=0.95\columnwidth]{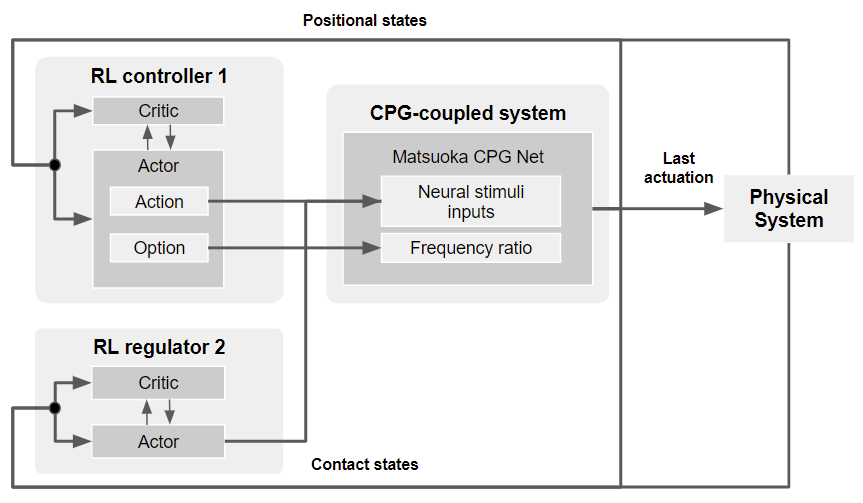}
    \caption{Schematic view of learning based CPG controller.}
    \label{fig:schematic_view}
\end{figure}

To this end, we develop a learning-based control method for obstacle-aided locomotion in soft snake robots. Using the limited perception of contact, this controlled robot will be aware of the existence of obstacles in the vicinity and react correctly by either taking obstacle-avoidance movement or obstacle-aided locomotion to achieve optimal goal tracking performance. This work is built on our previous work on soft snake robot \cite{xliu2020}, in which we established a model-free closed-loop goal tracking controller using central pattern generator (CPG) and reinforcement learning (RL). This controller is featured by a hierarchical framework where the RL controller learns to control the input to the CPG network, and the output of the CPG network will generate actuator inputs to the soft snake robot to achieve smooth and rhythmic locomotion in an obstacle-free environment. To enable reactive motions in a cluttered environment, we introduce a new control module--an event-triggered neural-network regulator, and design a game-theoretic approach that trains jointly the original RL controller and the new event-triggered regulator to achieve the optimal obstacle-aided locomotion.  Intuitively, we view the two controllers as two players in a cooperative game, with a shared goal to achieve successful and fast goal-tracking. 
Figure~\ref{fig:schematic_view} provides a schematic overview of the control design.   

Learning-based control for force interaction has recently been studied. 
The authors \cite{lee2020learning} have shown that a fully proprioceptive RL agent can learn four-legged locomotion on cluttered terrain without extra sensors by including additional  memory of history actions. Another work shows that a simulated ribbon eel could traverse the cluttered environment with RL controller learned in the free space\cite{min2019softcon}, under the assumption of perfect actuation without any noise or delay. However, both controllers do not incorporate the contact information into the feedback.  Our learning-based control design has the unique features of: 1) an event-triggered controller that reacts to contact force and the prediction of contact; 2) a combination of reinforcement learning and CPG to enable flexible and diverse locomotion patterns with smooth trajectories like biological systems; 3) a game-theoretic training process that achieves fast adaptation from controllers trained in an obstacle-free environment to two ``cooperative'' controllers in a cluttered environment.

The paper is organized as follows. First, we provide an overview of our schematic design of a soft snake robot equipped with contact sensors, as well as the state space representation in this work in Section~\ref{sec:sys}. In Section~\ref{sec:main}, we propose a learning-based control framework and game-theoretic training process. In Section~\ref{sec:experiment}, we design an experiment to showcase the improvement of the snake robot's locomotion performance in an obstacle-based environment. Section~\ref{sec:conclusion}  concludes and discuss future work.

\section{System Overview}
\label{sec:sys}

A full snake robot consists $n$ pneumatically actuated soft links. Each soft link of the robot is made of Ecoflex\texttrademark~00-30 silicone rubber. The links are connected through rigid bodies enclosing the electronic components that are necessary to control the snake robot. In addition, the rigid body components have a pair of one-direction wheels to model the anisotropic friction of real snakes. Only one chamber on each link is active (pressurized) at a time. 

In \cite{renato2019}, we developed a physics-based simulator that models the inflation and deflation of the air chamber and the resulting deformation of the soft bodies with tetrahedral finite elements. The simulator runs in real-time using GPU. Between the soft bodies, we have rigid bodies for mounting the electronic circuits, valves, and sensors. We use the simulator for generating data and learning the locomotion controller for the soft snake robot. The learned controller is then applied to the real snake robot. We have shown that with the high-fidelity simulator, the controller has a similar performance in both simulations and the real robot \cite{renato2019,xliu2020}.  

\begin{figure}[h!]
    \centering
    \includegraphics[width=0.7\columnwidth]{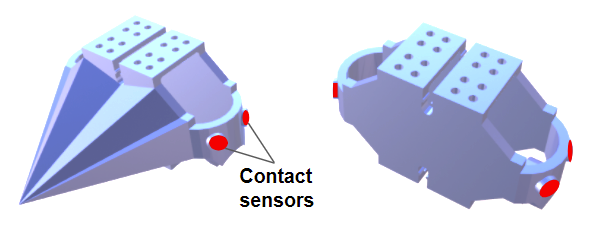}
    \caption{The 3D model of rigid head (left) and rigid body (right). The regions marked red are places installed with contact sensors.}
    \label{fig:sensor_view}
\end{figure}

\begin{figure}[h!]
    \centering
    \includegraphics[width=0.8\columnwidth]{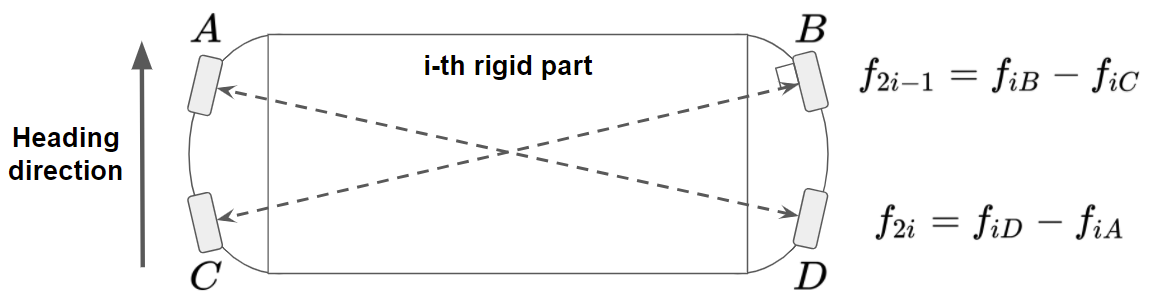}
    \caption{Free body contact force composition of $i$-th rigid body.}
    \label{fig:sensor_diagram}
\end{figure}

To enable force sensing and obstacle-aided locomotion, we add four point contact force sensors on two sides of each rigid body. The placement of these contact sensors is illustrated in Fig.~\ref{fig:sensor_view}. Inspired by a previous study on obstacle-aided locomotion of rigid snake robots \cite{liljeback2011experimental}, we approximate the total contact force acted on each rigid body based on the inputs sampled from the force sensors. As Fig.~\ref{fig:sensor_diagram} shows, each pair of sensors $A,D$ and $B,C$ are designed to have their diagonal line perpendicular to their sensing surfaces. This allows us to simplify the force representation by subtracting the two sensory inputs from a  pair of diagonal sensors. Therefore, for the $i$-th rigid body (counted from the head as $1$st rigid body), we have
\begin{align}
\label{eq:contact}
    &f_{2i-1} = f_{iB} - f_{iC},\\ \nonumber
    &f_{2i} = f_{iD} - f_{iA}.
\end{align}
For simplicity, we use a vector 
\[
    \mathbf{f} = [f_1, f_2, f_3, f_4, f_5, f_6, f_7, f_8, f_9, f_{10}]^T,
\]
to represent the vector of contact forces in a four-link soft snake robot with five rigid parts (including the head). This force representation  generalizes to $n$-link soft snakes, with $n\ge 4$.

\begin{figure}[h!]
    \centering
    \includegraphics[width=0.75\columnwidth]{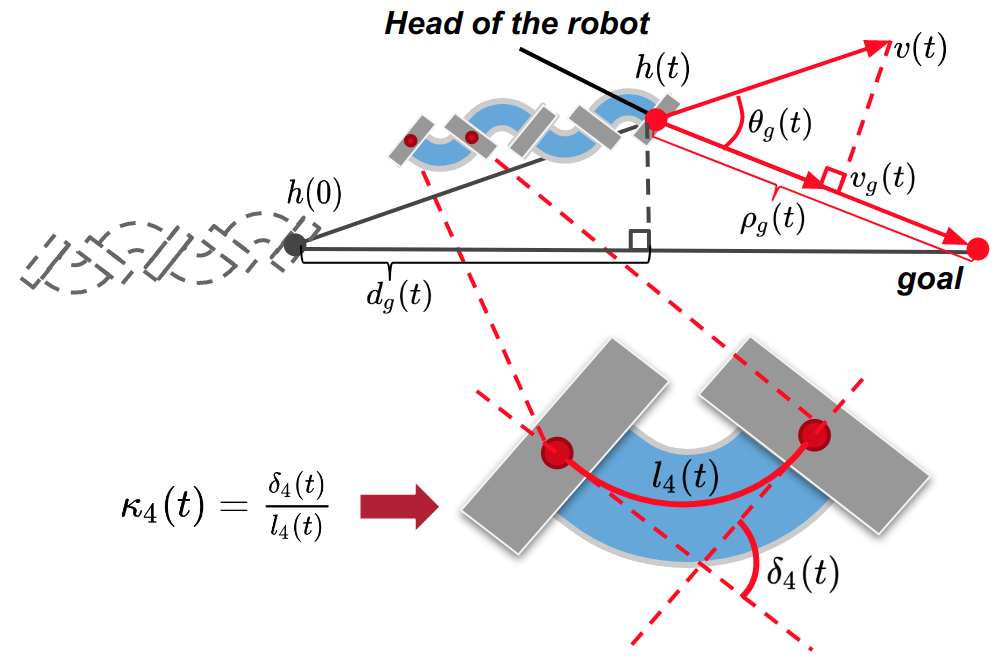}
    \caption{Notation of the state space configuration of the robot.}
    \label{fig:coordinate}
\end{figure}

The positional state space configuration (Fig.~\ref{fig:coordinate}) of the robot is the same as our previous setting in \cite{xliu2020}.

%
%
%
%
%
%
%
%
\section{Main Results}
\label{sec:main}

In this section, we describe our design of learning-based controller for obstacle-aided navigation with the soft snake robot. 
In section~\ref{sec:cpg}, we first review the \ac{cpg} system used for a learning-based control in obstacle-free environments. In section~\ref{sec:rl}, we present a \ac{rl} architecture to the locomotion control in a cluttered environment with multiple obstacles. We introduce two cooperative controllers, one is the original RL controller used in an obstacle-free environment, and another is an event-triggered RL regulator. Based on the theory and solution concepts of potential games, the proposed RL algorithm learns the Nash equilibrium using fictitious play.

\subsection{Learning-Based Controller with Matsuoka CPG Network}
\label{sec:cpg}

This section reviews the model-free RL method for soft-snake locomotion in  an obstacle-free environment, presented in our recent work \cite{xliu2020}. Our control design  employs
 Matsuoka oscillators as primitive nodes to construct a CPG network whose outputs translate to actuation inputs to  the soft snake robot.  

\noindent \textbf{Primitive Matsuoka CPG: }A primitive Matsuoka CPG consists a pair of mutually inhibited neuron models. The dynamical model of a primitive Matsuoka CPG is given as follows,
\begin{align}
	\begin{split}
		&K_f \tau_r \Dot{x}_i^e = -x_i^e - a z_i^f - b y_i^e - \sum_{j=1}^N w_{ji}y_j^e + u_i^e,\\ 
		&K_f \tau_a \Dot{y}_i^e = z_i^e - y_i^e,\\ 
		&K_f \tau_r \Dot{x}_i^f = -x_i^f - a z_i^e - b y_i^f - \sum_{j=1}^N w_{ji}y_j^f + u_i^f,\\ 
		&K_f \tau_a \Dot{y}_i^f = z_i^f - y_i^f,
	\end{split}\label{eq:matsuoka}
\end{align} 
where the subscripts $e$ and $f$  represent variables related to extensor neuron and flexor neuron, respectively. The tuple $(x_i^q, y_i^q)$, $q \in  \{e,f\}$ represents the activation state and self-inhibitory state of $i$-th neuron respectively,  $z_i^q = g(x_i^q) = \max(0, x_i^q)  $ is the output of $i$-th neuron,  $b \in \reals$ is a weight parameter,   $u_i^e, u_i^f$ are the tonic inputs to the oscillator, and  $K_f \in \reals$ is the frequency ratio.
The set of parameters in the system includes: the discharge rate $\tau_r \in \reals$, the adaptation rate $\tau_a \in \reals$,  the mutual inhibition weights between flexor and extensor $a\in \reals$ and the inhibition weight $w_{ji}\in \reals$ representing the coupling strength with the neighboring primitive oscillator. In our system, all coupled signals including $x_i^q, y_i^q$ and $z_i^q$ ($q \in  \{e,f\}$) are inhibiting signals (negatively weighted), and only the tonic inputs are activating signals (positively weighted).

In \cite{xliu2020}, we show that steering  control can be achieved by imbalanced tonic inputs. The steering bias of a primitive Matsuoka oscillator is proportional to the amplitude of $u^e$ when $u^e$ and $u^f$ are exclusive within the range $[0, 1]$, that is $u^f =1-u^e$. 
As the tonic inputs have to be positive in Matsuoka oscillators,  we define a four dimensional action vector $\vec{a}= [a_1, a_2, a_3, a_4]^T \in \reals^4$ and map $\vec{a}$ to tonic input vector $\vec{u}$ as follows, 
\begin{equation}
	\label{eq:decoder}
	u_i^e = \frac{1}{1+e^{-a_i}},\text{ and } 
	u_i^f =1-u_i^e, \text{ for } i=1,\ldots, 4.
\end{equation}
This mapping bounds the tonic input within $[0, 1]$.  The input vector $\mathbf{u}$ for the 4-link snake robot  is an eight-dimension vector, 
\begin{equation*}
    \mathbf{u} = [u_1^e, u_1^f, u_2^e, u_2^f, u_3^e, u_3^f, u_4^e, u_4^f]^T.
\end{equation*}
We use the notation $\mathbf{u^e}$ or $\mathbf{u^f}$ to represent the vector of all extensor or flexor elements of $\mathbf{u}$ respectively, i.e.,
\[
    \mathbf{u^e} = [u_1^e, u_2^e, u_3^e, u_4^e]^T, \quad \mathbf{u^f} = [u_1^f, u_2^f, u_3^f, u_4^f]^T.
\]
From our previous work \cite{xliu2020}, we have discovered that the  velocity control of the snake robot can be achieved by the frequency ratio $K-f$-a unique feature of Matsuoka CPG networks. Thus, we determine the encoded input vector of the \ac{cpg} net to be tonic input vector $\vec{u}$ for steering control and frequency ratio $K_f$ for velocity control. 

This input vector of the \ac{cpg} is the output vector of the \ac{nn} controller. The input to the \ac{nn} controller is the state feedback of the robot, given by $\mathbf{s}=[\rho_g, \Dot{\rho_g}, \theta_g, \Dot{\theta_g}, \kappa_1, \kappa_2, \kappa_3, \kappa_4]^T \in \reals^8$. 

We name the \ac{nn} controller as the RL controller 1 (C1)  and the proposed scheme that connects R1 and CPG net as a PPOC-CPG \cite{xliu2020}. The RL controller is trained with the option-critic method \cite{bacon2017option} to  control both tonic inputs $\vec{u}$ and frequency ratio $K_f$ in the CPG system. 

\subsection{Event-triggered RL regulator}
\label{sec:rl}

The PPOC-CPG controller learned in obstacle-free environments does not take contact sensory information as its feedback to modulate the control of the soft snake robot. To enable contact-awareness and force feedback in closed-loop control, we introduce an event-triggered controller, called R2. An event-triggered controller only outputs actuator signals when an event-triggering condition is satisfied.  

First, we define the \emph{event-triggering condition} as follows: 
At each time step, given the contact force vector $\mathbf{f}$ and the distance $d_o \in \reals$ between the head of the snake robot and the nearest obstacle. Define $D\in \reals$ as a detection threshold that limits the spatial sensing range of the robot. The event-triggering condition for the contact-aware scenario is $\norm{f} \neq 0$ or $d_o < D$. When the event-triggering condition is satisfied, the contact-aware regulator is triggered to join the manipulation of the CPG system. 

In the obstacle-based locomotion scenario, there are in total $26$ observation states $\mathbf{\zeta} = \{\zeta_1, \zeta_2, ..., \zeta_{26}\}$, where $\zeta_{1:4}$ represents the dynamic state of the robot referenced on the goal position, $\zeta_{5:8}$ represents the real-time body curvature of the $4$ soft links, $\zeta_{9:14}$ contains actions in the last time step including the previous option and the terminating probability, $\zeta_{15:24}$ contains the pre-processed contact forces, and $\zeta_{25:26}$ are the distance and relative angle to the closest obstacle in the environment. 

The regulator is also a \ac{nn} controller that takes the total observation states and is trained using the actor-critic method. Specifically, the RL regulator and the RL controller (C1) share the control of the tonic inputs to the CPG system, as described next. However, only the RL controller  can control the frequency ratio of the CPG.

\noindent \textbf{Linear tonic input composition between C1 and R2:} According to the bias property of the Matsuoka oscillator\cite{xliu2020}, let $\Delta u_i = u_i^e - u_i^f$, it can be shown that
\[
bias(\Delta u_i) \propto bias(\psi_i),
\]
where $\psi_i$ is the output wave of the Matsuoka oscillator. That is, the bias between the extensor and flexor tonic inputs is linearly proportional to the oscillation bias of the output wave of the Matsuoka oscillator. This property implies that a linear composition of tonic input signals generated from different sources  yields a linear composition of oscillation bias, regardless of the oscillation feature of each source signal. Inspired by this result, the event-triggered RL regulator can  manipulate the bias of tonic input through direct linear superposition to the control signal generated by the RL controller. 

Let $\mathbf{a}_1$  represent the action vector generated by the controller (C1) and $\mathbf{a}_2$   represent the action vector generated by the regulator respectively. According to \eqref{eq:decoder}, the raw actions needs to be further decoded into tonic input vectors. Thus we use $\mathbf{u_k}$ to represent the tonic inputs generated from $\mathbf{a_k}$, $k=1, 2$. We have the final tonic input for the \ac{cpg} network as a linear composition of the tonic inputs by C1 and R1, 

\begin{equation}
    \label{linearcompos}
    \mathbf{u} = w_1 \mathbf{u_1} + w_2 \mathbf{u_2},
\end{equation}
where $w_1$ and $w_2$ are positive constant weights satisfying $w_1 + w_2 = 1$, and $\mathbf{u}$ is the vector representing total tonic inputs to the CPG network\cite{xliu2020}. 

\normalcolor
\subsection{Fictitious cooperative game for learning-based control}

The learning-based control design in an obstacle-free environment \cite{xliu2020} is formulated as a reinforcement learning in a Markov decision process, which is a single-agent decision-making problem. In obstacle-aided locomotion, by introducing the event-triggered RL regulator, we propose to model the interaction between the robot and its environment/obstacles as a \emph{two-player cooperative} game $G= (S, A_1\cup A_2,P, R,\gamma)$
where $S$ is the set of states, $A_1$ is the action set of player 1(RL controller, C1) and $A_2$ is the action set of player 2 (RL regulator, R2). The transition function $P: S\times A_1\times A_2\rightarrow \dist{S}$ is unknown. The two players share the same reward function $R: S\times A_1\times A_2\rightarrow \reals $ and $\gamma \in (0,1)$ is the discounting factor.  
Specifically,  the two inputs from C1 and R2 are combined through a weighted sum \eqref{linearcompos}. 

In this cooperative game, players share the common reward function and aim to maximize the discounted sum of rewards. Let $\pi_i$ be the strategy of player $i$, for $i=1,2$. The payoff of player $i$ given a strategy profile $(\pi_1,\pi_2)$ and the initial state $s\in S$ is 
\[
V_i(s,\pi_1,\pi_2) = \sum_{t=0}^\infty \gamma^t \Expect(R_t\mid \pi_1,\pi_2, s_0=s).
\]
where $R_t$ is the reward obtained at the $t$-th step in the stochastic process induced by the strategy profile.

Because the reward function is common to both players, we have $V_1(s,\pi_1,\pi_2) = V_2(s,\pi_1,\pi_2)$ for all $s\in S$ and all possible strategy profiles. The cooperative game is a special case of potential game where the payoff function $V_i$ equals the potential function $V$. 

The Nash equilibrium is a tuple $(\pi_1^\ast,\pi_2^\ast )$ such that 
\[
V_i(s,\pi_i^\ast, \pi_j^\ast) \ge V_i(s,\pi_i, \pi_j^\ast), \text{ for } (i,j)\in \{(1,2), (2,1)\}.
\]

To enable learning in the cooperative game, we employ fictitious play. First, we introduce the notion of perturbed best-response \cite{hofbauerGlobalConvergenceStochastic2002} strategy for player $i$ given player $j$'s strategy $\pi_j$, which is 
\begin{align*}
    \pi_i^{BR}(s,a ) \propto \exp(\lambda  R(s,a, \pi_j(s)) + &\Expect_{s'}\left[ V_i(s', \pi^{BR}_i, \pi_j) \right]), \\
    &\forall s\in S, \forall a \in A_i,
\end{align*}
where $\lambda >0$ is a pre-defined constant and $\propto$ is the symbol for ``proportional to''.

 The perturbed best-response strategy for player $i$ is the (near-) optimal policy in the Markov decision process  obtained from the game $G$ by marginalizing out the other player's strategy. 
 
 Next, we describe our design of smooth fictitious play to learn the Nash equilibrium in the Markov game: At the round $0$, we initialize the RL controller to be the trained goal-tracking controller $\pi_1^0$ in an obstacle-free environment and fix the RL controller. We then train the RL regulator until it converges to the perturbed best-response strategy. Let $n$ denote the inner episode of the RL algorithm and $V_i^n(s)$ is the reward acquired at episode $n$ in the RL learning process. Upon the convergence of episode reward ($||V_i^n(s)-V_i^{n-1}(s)|| \leq \epsilon$) measured by the RL algorithm (PPO Learning), a new round starts and the RL controller learns to best-respond to the current strategy of RL regulator. This process iterates for a number of rounds until both controllers converge.
 
 It is noted that we employ the softmax instead of hardmax in the computation of the perturbed best response. This is needed to avoid the sensitivity issues in fictitious play. Besides,  the response of one player is calculated given the knowledge of the opponent's strategy instead of the empirical estimate of the opponent's strategy. Our approach is a class of multi-agent RL, in which each agent employs RL algorithms treating other agents as part of the probabilistic environment. Due to the potential game formulation, the convergence to the Nash equilibrium is ensured using the fictitious play \cite{hofbauerGlobalConvergenceStochastic2002}.
 


Finally, the learning algorithm is concluded in Algorithm~\ref{alg:fictitious-play} below. $\pi_1^0$ is the policy of C1 which has been initially trained in an obstacle-free environment, while $\pi_2^0$ is a  random policy for R2 at the beginning. The two policies are first evaluated in the obstacle-based training environment in order to initialize the  value function $V^0$. We use a flag to determine which players should be updated at the current iteration. In the algorithm, the updating flag switches when the actor-critic learning process exceeds a manually set terminating time. For each macro iteration step $i$, only one of the players is updated, while the other one has all its parameters frozen. The algorithm terminates until the improvement of the value function in the new iteration is below a certain threshold $\epsilon$ or an upper bound on the total number of iterations is reached.





\begin{algorithm}
 	\SetAlgoLined
	\KwResult{Return a (near-)optimal policy pair: $\pi_1^N$ for the controller (C1) and $\pi_2^N$ for the regulator (R2).}
	\KwInput{An initial policy $\pi_1^0$ for C1 and $\pi_2^0$ for R2, two threshold constant $\epsilon_i,i=1,2$, an upper bound $N$ on the number of iterations }
	\SetKwFunction{Evaluation}{PolicyEval}
	\SetKwFunction{RL}{PPO Learning}
	$V^0 \leftarrow \Evaluation(\pi_1^0,\pi_2^0)$;\Comment{Simulation-based policy evaluation.}\\
	\Comment{The superscript is the number of the macro iteration.}
	$\textsc{flag} \leftarrow $ R2\;
	\For{$ i =  1, \ldots, N$}{
		\eIf{$\textsc{flag}=$C1}{
			$ (V^i, \pi_1^i)  \leftarrow \RL(G,  \pi_2^{i-1})$; \Comment{Fix the policy for R2, and employ RL to learn the controller C1}\\
		$\textsc{flag}\leftarrow$  R2\;}{
			$(V^i, \pi_2^i)  \leftarrow \RL(G,  \pi_1^{i-1})$; \Comment{Fix the policy for C1, and employ RL to learn the regulator R2}\\
				$\textsc{flag}\leftarrow$  C1\;}
    	\If{$\norm{V^i(s_0) -V^{i-1}(s_0)} \le \epsilon $}{
            \Return $(\pi_1^i, \pi_2^i)$\;
        }
	}
	\caption{Fictitious cooperative game to train the controller and the regulator.}
	\label{alg:fictitious-play}	
\end{algorithm}


\subsection{Design of the shared reward function}
\label{sec:reward}

\begin{figure}[h!]
    \centering
    \includegraphics[width=0.3\columnwidth]{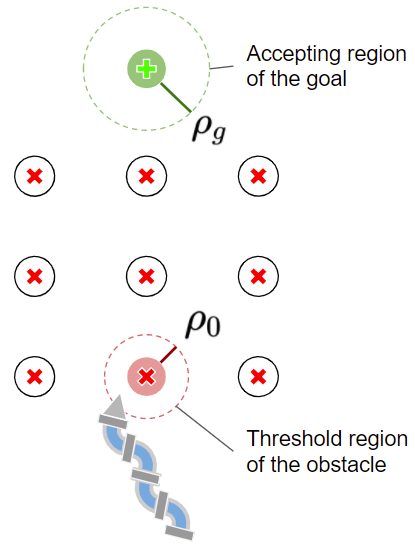}
    \caption{Illustration of obstacle-aware locomotion scenario with goal reaching tasks.}
    \label{fig:scenario}
\end{figure}

We now present our design for the reward function shared by C1 and R2. Our design will ensure that by maximizing the discounted sum of reward, the learned controller can achieve efficient locomotion and accurate set-point tracking. 

To improve learning efficiency, we employ a potential field-based reward function. Artificial potential field (APF) is widely applied in planning problems and potential game theory \cite{khatib1986real, park2001obstacle, dong2012strategies} to accelerate the process of searching for the optimal strategy. 

In this paper, we use a simple APF to formulate the shared reward function for both the controller agent and the regulator agent. The potential field can be classified into two categories -- the attracting field for target reaching  and the repulsive field for obstacle avoidance. The attracting field function is defined as follows
\[
    U_{att}(\mathbf{p}) = \frac{1}{2} k_{att} ||\mathbf{p} - \mathbf{p}_g||^2,
\]
where $\mathbf{p}$ is the coordinate of the agent and $\mathbf{p}_g$ is the coordinate of the goal. Coefficient $k_{att}$ is a positive constant indicating the strength of the potential field. Since the attracting gravity is always pointing toward the goal coordinate from any position of the map, the value of gravity force should be negative. By taking the negative gradient of $U_{att}$, we have the attracting force function
\[
    \mathbf{F}_{att}(\mathbf{p}) = -\nabla U_{att} = -k_{att} (\mathbf{p}-\mathbf{p}_g).
\]

For obstacle avoidance purposes, we adapt the repulsive potential function named FIRAS defined in\cite{khatib1986real}. The repulsive potential function of the $i$-th obstacle in the environment is represented as
\[
    U_{rep}(\mathbf{p};i) = \begin{cases}
        \frac{1}{2}k_{rep}(\frac{1}{\rho_i} - \frac{1}{\rho_0})^2, \quad \textit{if } \rho_i \leq \rho_0\\
        0 \quad \textit{if } \rho_i > \rho_0.
    \end{cases}
\]
where $\rho_i = ||\mathbf{p} - \mathbf{p}_{o_i}||$ indicates the shortest distance from the robot to the $i$-th obstacle, and $\rho_0$ is a threshold constant indicating that the repulsive field is only effective in the vicinity of an obstacle. 

When there are multiple obstacles, the total potential force is supposed to be the summation of all potential forces generated by all obstacles involved in the task,
\begin{align*}
     \mathbf{F}_{rep}(\mathbf{p}) &= \sum_{i = 0}^{n} F_{rep}(\mathbf{p};i) = \sum_{i = 0}^{n} \nabla U_{rep}(\mathbf{p};i) \\
     &= \sum_{i = 0}^{n} k_{rep} (\mathbf{p} - \mathbf{p}_{o_i}) \frac{1/\rho_i - 1/\rho_0}{\rho_i^3}.
\end{align*}

The reward is designed to encourage the goal-reaching and following the guide by the artificial potential field. We design the reward to be composed of three rewards:  
\begin{equation}
	R = \omega_1 R_{goal}  + \omega_2 R_{att}  + \omega_2 R_{rep},
\end{equation}
where $\omega_i,i=1,2,3$ are constant weights. $R_{goal}$ is the termination reward for reaching a circular accepting area centered at the goal. 
\[
R_{goal}= \cos \theta_g \sum_{k=0}^i{\frac{1}{l_k} \mathbf{1}(\rho_g < l_k)}.
\]
where $\theta_g$ is the deviation angle between the locomotion direction of the snake robot and the direction of the goal, $l_k$ defines the radius of the accepting area in task-level $k$, for $k=0,\ldots, i$. $\rho_g = \norm{ \mathbf{p}-\mathbf{p}_g}$ is the linear distance between the head of the robot and the goal, and $\mathbf{1}(\rho_g < l_k)$ is an indicator function to determine whether the robot's head is within the accepting area of the goal. 
$R_{att}$ (resp. $R_{rep}$) is the reward function of the attracting potential field (resp. repulsive potential field):
\[
R_{att} = \mathbf{v} \cdot \mathbf{F}_{att}(\mathbf{p}), \quad
R_{rep} = \mathbf{v} \cdot \mathbf{F}_{rep}(\mathbf{p}),
\]
where $\mathbf{v}$ is the velocity vector. The dot product between $\mathbf{v}$ and the potential field vector represents the extent of the agent's movement on following the potential-flow in the task space. In this reward design, though the repulsive potential function generates a cost for the contact, its combination with the other two reward terms may encourage the contact especially when the contact force can aid the locomotion. We discuss this aspect in the experimental validation.

\section{Experiments}
\label{sec:experiment}
\subsection{Basic Settings}

 We use a 4-layer neural network with $128\times128$ hidden layer neurons for both controller (C1) and regulator (R2). For the total $26$ observation states $\mathbf{\zeta} = \{\zeta_1, \zeta_2, ..., \zeta_{26}\}$, C1 only takes $\zeta_{1:14}$ as the inputs to the policy network, while R2 takes all observation states $\zeta_{1:26}$ as its inputs. The outputs for C1 includes an action vector $\vec{a}_1$ to control the tonic inputs of the \ac{cpg} network, and an option vector $\mathbf{o} = [o, \beta]^T$ to control the change of frequency parameter of the neural oscillator. The outputs for R2 contain only the action vector $\vec{a}_2$ adding to the tonic inputs. The hyper-parameters for the policy network and value network in both C1 and R2 are the same as our previous work\cite{xliu2020}. 


\paragraph{Task specification} As Fig.~\ref{fig:scenario} shows, the robot is required to traverse a group of obstacles and reach the goal sampled in a certain range of distances and deviation angles from the initial position and heading of the robot. The dotted circle around the goal indicates the accepting area, which means that the robot needs to be close enough to the goal in order to succeed and receive a terminal reward. At each time step, the robot also receives a reward from the potential field defined in Section~\ref{sec:reward}. If the agent reaches the accepting region of the current goal, a new goal is randomly sampled. There are two failing situations, where the desired goal will be re-sampled and updated. The first situation is starving, which happens when the robot is jammed for a certain amount of time. The starvation time for failing condition is $900 \text{ ms}$. The second case is missing the goal, which happens when the robot keeps heading a wrong direction as oppose to moving towards the goal for a certain amount of time.  Specifically, in our experiment, if the linear velocity of the snake robot stays negative on the goal direction for over $60$ time steps, then we consider that the robot misses the goal.

\paragraph{Training Environment} Figure~\ref{fig:scenario} shows a snapshot of the training environment with a $3\times 3$ obstacle maze for each goal reaching task. The distance between the robot and the goal is fixed to $1.5$ meters. The deviation angle between the snake robot and the goal is initially sampled from $0$ degree to $60$ degrees with a uniform distribution. The base distance between every two obstacles is $0.08$ meters. The coordinate of each obstacle is added by an additional clipped standard Gaussian noise ($\omega \sim \mathcal{N}(0, 1)$, clipped by $-0.01<\omega<0.01$).

\paragraph{Fictitious play results} 
\begin{figure}[h!]
	\centering
	\includegraphics[width=0.8\columnwidth]{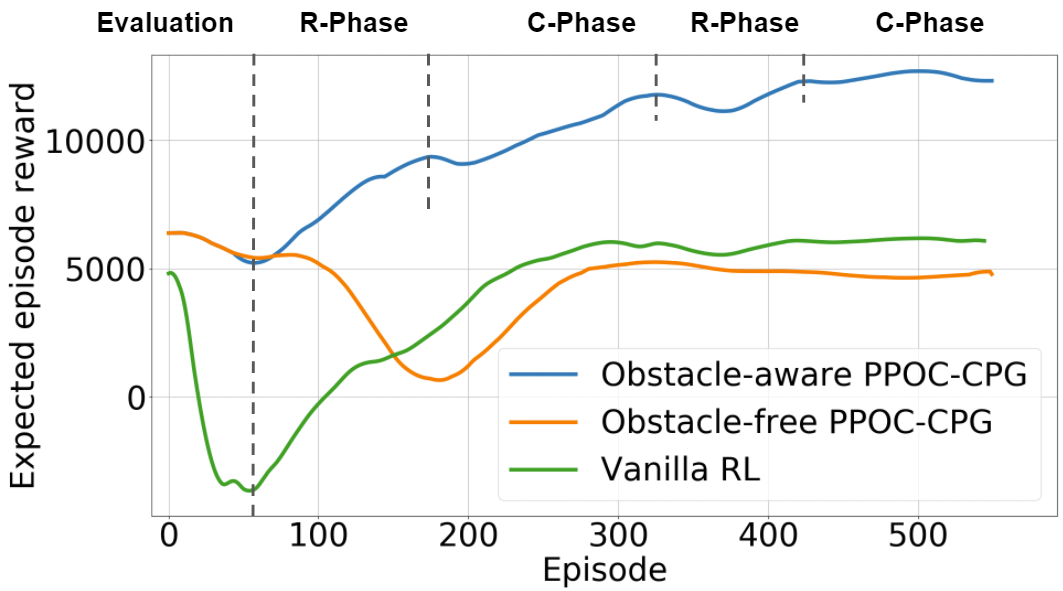}
	\caption{Learning process of fictitious cooperative game with PPOC-CPG compared with vanilla PPOC-CPG, and vanilla PPO, all three methods are recorded in an obstacle-based training environment.}
	\label{fig:compare_result}
\end{figure}
Figure~\ref{fig:compare_result} illustrates the learning process of Alg.~\ref{alg:fictitious-play} applied to the obstacle-aware locomotion learning of the soft snake robot. First, the initial policies $\pi_1^0, \pi_2^0$ for C1 and R2 are evaluated in the obstacle-based environment, during which the parameters of both $\pi_1^0$ and $\pi_2^0$ are frozen. Particularly, $\pi_1^0$ has its policy network pre-trained in an obstacle-free environment\cite{xliu2020}. This step is to obtain a proper estimation of the approximated value function of the initial method in the new environment. After that, the actor-critic algorithm (PPO) only updates R2's parameters in macro iteration $1$, which is the first "R-phase" in Fig.~\ref{fig:compare_result}. Then the algorithm iterates until the difference of value function between two consecutive iterations drop below $\epsilon = 10$ in our experiment setting. It is noticed that every time after the switch, the expected episode reward decreases slightly. This is because the learning player's initial strategy has not converged to the best response given the new environment (due to the change in the policy of the other player).

Besides, we compare our proposed controller with a single PPOC-CPG that trains $\pi_1$ of the controller in the obstacle-based environment. It is observed that the expected episode reward keeps dropping after the evaluation phase due to the exploration in the new environment. Then the reward recovers to the value that is close to the end of the evaluation phase, but never has a chance to keep improving and surpass the initial performance. This shows the significance of the regulator on improving the overall performance in the environment with multiple obstacles.



\begin{table*}[t]
    \centering
    \vspace{0.02\columnwidth}
    \caption{Performance Comparison}
    \scalebox{1.0}{
    \begin{tabular}{c|c|c|c|c}
        \hline
        \textbf{Method} & \textbf{Jam ratio} & \textbf{Avg. linear velocity } ($m/s$) & \textbf{Success rate} & \textbf{Avg. time per goal} ($s$)\\
        \hline
        obstacle-aware PPOC-CPG & \textbf{0.165} & \textbf{0.139} & \textbf{0.91} & \textbf{36.1} \\
        obstacle-free PPOC-CPG & 0.235 & 0.124 & 0.82 & 40.8 \\
        Vanilla PPO & 0.279 & 0.093 & 0.74 & 61.7  \\
        \hline
    \end{tabular}}
    \label{tab:comparison}
\end{table*}

\subsection{Performance evaluation and comparison analysis }

\begin{figure}[h!]
    \centering
    \includegraphics[width=0.8\columnwidth]{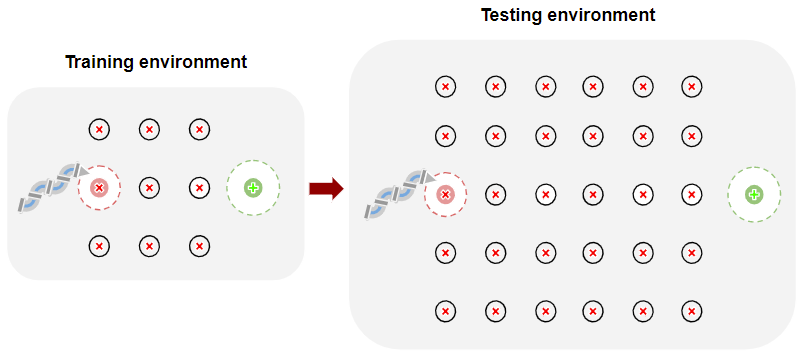}
    \caption{Obstacle configuration of training and testing environments.}
    \label{fig:traintotest}
    \vspace{-1ex}
\end{figure}




\begin{figure*}[h!]
\centering
\vspace{0.02\columnwidth}
\begin{subfigure}[b]{0.4\columnwidth} 
\includegraphics[width=\columnwidth]{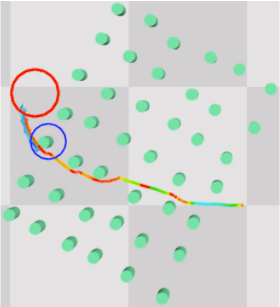}
\caption{$0$ episode}
\label{fig:epia}
\end{subfigure}
\hspace{0.01\columnwidth}
\begin{subfigure}[b]{0.4\columnwidth} 
\includegraphics[width=\columnwidth]{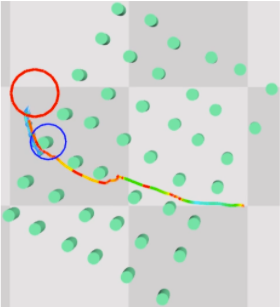}
\caption{$100$ episode}
\label{fig:epib}
\end{subfigure}
\hspace{0.01\columnwidth}
\begin{subfigure}[b]{0.4\columnwidth} 
\includegraphics[width=\columnwidth]{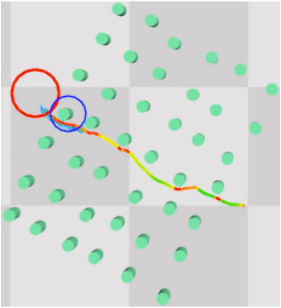}
\caption{$500$ episode}
\label{fig:epic}
\end{subfigure}
\hspace{0.01\columnwidth}
\begin{subfigure}[b]{0.425\columnwidth} 
\includegraphics[width=\columnwidth]{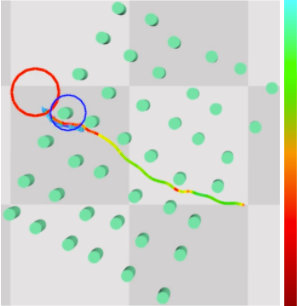}
\caption{$1000$ episode}
\label{fig:epid}
\end{subfigure}
\caption{Sample navigation paths controlled by the policy of the obstacle-aware PPOC-CPG tested at different learning stage. The color bar on the right indicates the value of potential reward gained at each time step. }
\vspace{-2ex}
\label{fig:compareiters}
\end{figure*}

\begin{figure*}[h!]
\centering

\begin{subfigure}[b]{0.69\columnwidth} 
\includegraphics[width=\columnwidth]{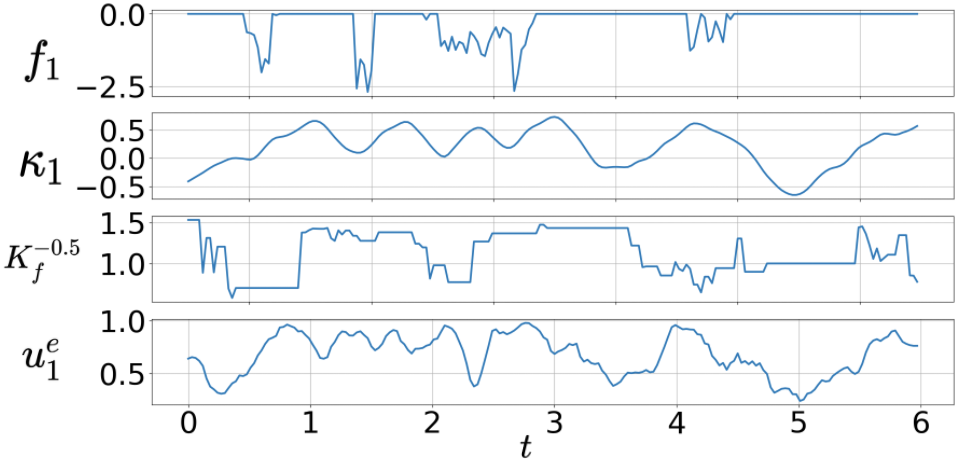}
\caption{}
\label{fig:negevent}
\end{subfigure}
\begin{subfigure}[b]{0.65\columnwidth} 
\includegraphics[width=\columnwidth]{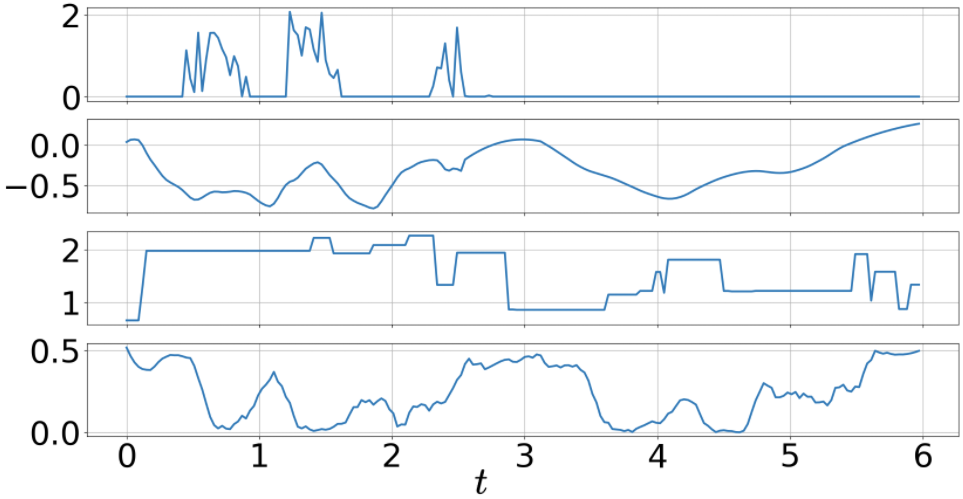}
\caption{}
\label{fig:posevent}
\end{subfigure}
\begin{subfigure}[b]{0.65\columnwidth} 
\includegraphics[width=\columnwidth]{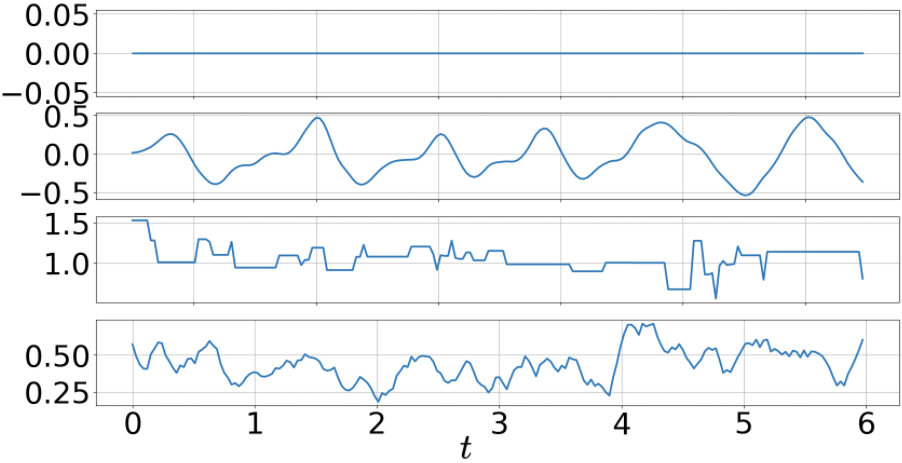}
\caption{}
\label{fig:normevent}
\end{subfigure}

\caption{Sample action trajectories of obstacle-aware PPOC-CPG controller in the situations with (a) negative contact force (b) positive contact force and (c) zero contact force value of the first contact sensor $f_1$. Note $\kappa_1$ is the curvature of the first link. }
\vspace{-2ex}
\label{fig:compareevents}
\end{figure*}






To demonstrate  the effectiveness of the proposed scheme, we compare the performance in obstacle-aware locomotion for three different learning-based controllers: (1) the PPOC-CPG equipped with contact-aware regulator (obstacle-aware PPOC-CPG, proposed); (2) the PPOC-CPG trained in free space; and (3) the vanilla PPO (a neural network without CPG) trained in the environment without obstacles. All controllers are trained in the training environment in Fig.~\ref{fig:traintotest} until convergence. 

The data for comparison is collected from a testing scenario which is more complicated than the training environment. In the testing environment, there are more number of obstacles in each task, and the coordinate of each obstacle is also added by a standard Gaussian noise ($\omega \sim \mathcal{N}(0, 1)$, clipped by $-0.01<\omega<0.01$). In the testing case, the distance between the robot and the goal is fixed to $2$ meters, and the initial deviation angle between the snake robot and the goal is uniformly sampled from $0$ degree to $90$ degrees. As Fig.\ref{fig:traintotest} shows, each task in the testing scenario includes  $5\times 6$ identical circular obstacles (radius=$0.02$ meters) between the initial position of the robot and the goal. 
 In addition to total accumulated reward, we also evaluate each controller's capability of dealing with the contact situation.  Since we do not assume that the head of the robot could always avoid obstacles perfectly, we add a jam-detection module such that the entire snake robot is also determined to be jammed if its head joint is jammed and its linear velocity stays lower than a threshold over a certain amount of time. In all experiments of this paper we set this low velocity threshold to be $v_0 = 0.02$ m/s and the time threshold to be $300$ ms. By calculating the total time of the snake robot being jammed throughout the testing process, we further adapt this statement to a measurement named "jam ratio", such that 
\[
jam\text{ }ratio = \frac{total\text{ }jam\text{ }time}{total\text{ }task\text{ } time}.
\]

We run the soft snake robot with three different controllers in the testing scenario for $100$ tests each and compare the performance with several measurement criteria listed in Table~\ref{tab:comparison}. For the calculation of the success rate in the testing scenario, the starvation criteria also applies. The results in Table~\ref{tab:comparison} show  that the obstacle-aware PPOC-CPG controller outperforms the other two significantly, with the lowest jam ratio, highest average linear velocity and highest success rate. This result provides an important evidence on the capability of the obstacle-aware PPOC-CPG on improving the efficiency of obstacle-aware locomotion. In addition, it is worth noticing that the PPOC-CPG trained in free space also shows good adaptability to the obstacle-based environment with memoryless and fully proprioceptive observations. 
To better illustrate the improvement of obstacle-aware locomotion strategy progressively, we retrain the model using Alg.~\ref{alg:fictitious-play} for $1000$ actor-critic episodes and test the control policy of obstacle-aware PPOC-CPG at different learning stage in a goal-reaching task. The resulting locomotion traces are presented in Fig.~\ref{fig:compareiters}. The colors on the paths represent the potential reward at each segment of a path. The lighter the color, the higher the reward as indicated by the color bar on the right of the figure. Besides, the dark-colored segments in the path usually indicate a notable contact when they are close to the obstacles. Such contact could slow down the robot or even cause a jam. From the figure, we can observe a significant improvement of the potential reward collected by the snake robot along with the learning process. Particularly, the policy switches the preferred path between $100$ episodes and $500$ episodes, and significantly reduces the portion of low potential reward by either avoiding the contact or taking the contact more efficiently. 




Next we plot three typical trajectories of the first link actions when the $f_1$ sensor on the head link detects negative contact force ($f_1<0$), positive contact force ($f_1>0$) and no contact force based on Eq.~\eqref{eq:contact}. The contact force is defined in Fig.~\ref{fig:sensor_diagram}. In Fig.~\ref{fig:negevent}, there are three consecutive force signals and one minimal force detected by the sensors. From $t = 0$ s to $t=3$ s, the tonic input signal as a linear composition of transformed actions by C1 and R2 shows the same trend of amplifying the extensor tonic input, which increases the oscillating bias to a positive level. In the meantime, the inverse of frequency parameter $K_f^{-0.5}$ also increases, which indicates an increase of oscillation frequency given that $\hat{\omega} \propto \frac{1}{\sqrt{K_f}}$ \cite[Eq.(11)]{xliu2020}. These actions tend to cause more contacts to create enough lateral propulsion force that pushes the snake robot away from the obstacle. Right after the contact vanishes at around $t=3.1$ s, the regulator R2 reacts quickly and reduces the extensor tonic input of the head joint to zero to recover normal oscillation, with the oscillation frequency recovered at the same time. Figure~\ref{fig:posevent} roughly shows an anti-symmetric behavior to the trajectories in Fig.~\ref{fig:negevent} when the contacts occur on the other side of the head link. A more significant difference can be observed by further comparing the contact-aware trajectories with a plot of contact-free trajectories in Fig.~\ref{fig:normevent}, where the oscillation signal shows a relatively more steady oscillation with zero bias and stabilized frequency. 
  
\section{Conclusion}
\label{sec:conclusion}

This paper develops a learning-based controller for obstacle-aware locomotion with a soft snake robot. The control scheme employs two cooperative controllers: one RL controller is trained with free space locomotion and adapted to interact with obstacles, while the other RL regulator is event-triggered and cooperates with the RL controller to improving the locomotion efficiency while the robot interacts with and navigates through obstacles to achieve the desired goal position. Both controllers 
manipulates the inputs to a Matsuoka CPG system, whose outputs translate to the actuation inputs for the soft snake robot. 
 As a dynamical system, the special property of the Matsuoka CPG system enables us to design a linear composition of the outputs of the two modules for steering control, and also enables the separation of steering and velocity control. This allows the soft snake robot to adapt to the obstacle-based environment rapidly given its previous knowledge for locomotion in an obstacle-free environment. The simulation result shows a significant improvement in locomotion performance in a cluttered environment. In the future, we plan to carry out extensive study in the physical system and investigate the relation between the actions of C1 and R2 and the placement of the contact sensors. Moreover, it is hypothesized that the link number of the snake robot is too few to realize more efficient obstacle-aided locomotion. We plan to model and manufacture a longer soft snake robot ($\ge 10$ links). We expect a longer snake could create more contacts between its body and the environment and leverage some of these contact points on its body as footholds to propel it towards the desired direction for locomotion efficiency.



\bibliographystyle{IEEEtran}
\bibliography{refs.bib}
\vspace{4em}


\end{document}

%% file: defs.tex
\newif\ifuseboldmathops
\newif\ifuseittextabbrevs
\useboldmathopstrue   

\ifuseittextabbrevs

	\newcommand{\etal}{{et~al.}}
\else

	\newcommand{\etal}{et~al.}
\fi

\ifuseboldmathops
	\newcommand{\reals}{\mathbf{R}}

\else
	\newcommand{\reals}{\mathbb{R}}

\fi

\ifuseboldmathops

\else

\fi

\ifuseboldmathops
	\newcommand{\Expect}{\mathop{\bf E{}}\nolimits}
	
\else
	\newcommand{\Expect}{\mathop{\mathbb{E}{}}\nolimits}
	
\fi

\ifuseboldmathops

\else

\fi

\newcommand{\norm}[1]{\lVert#1\rVert}

\newcommand{\dist}[1]{\Delta(#1)}

\renewcommand{\vec}[1]{\mathbf{#1}}

\acrodef{mdp}[MDP]{Markov Decision Process}
\acrodef{pomdp}[POMDP]{Partially Observable Markov Decision Process}

\theoremstyle{definition}

\acrodef{smdp}[Semi-MDP]{Semi-Markov decision process}
\acrodef{rl}[RL]{reinforcement learning}
\acrodef{mcts}[MCTS]{Monte Carlo tree search}
\acrodef{uct}[UCT]{Upper Confidence Bound 1 applied to trees}
\acrodef{scltl}[scLTL]{syntactically co-safe LTL}
\acrodef{ssp}[SSP]{Stochastic Shortest Path}
\acrodef{p2sg}[SG(2)]{Two-player Stochastic Game}
\acrodef{dof}[DOF]{degree of freedom}
\acrodef{cpg}[CPG]{Central Pattern Generator}
\acrodef{nn}[NN]{Neural Network}
\acrodef{snn}[SNN]{Spiking Neural Net}
\acrodef{rstdp}[R-STDP]{Reward-Modulated Spike-Timing-Dependent Plasticity}
\acrodef{gp}[GP]{Genetic Programming}
\acrodef{ppoc}[PPOC]{Proximal Policy Optimization Option-Critics}
\acrodef{dr}[DR]{Domain Randomization}
\acrodef{bibo}[BIBO]{Bounded-input, Bounded-Output}

\acrodefplural{dof}[DOFs]{degrees of freedom}
\acrodefplural{cpg}[CPGs]{Central Pattern Generators}
